\documentclass{article}
\usepackage{ijcai19}

\usepackage{times}
\usepackage{soul}
\usepackage{url}
\usepackage[hidelinks]{hyperref}
\usepackage[utf8]{inputenc}
\usepackage[small]{caption}
\usepackage{graphicx}
\usepackage{amsmath}
\usepackage{booktabs}
\usepackage{algorithm}
\usepackage{algorithmic}
\title{Learning High-Level Planning Symbols from Intrinsically Motivated Experience}

\author{
Angelo Oddi\and
Riccardo Rasconi\and
Emilio Cartoni\and
Gabriele Sartor\and
Gianluca Baldassarre\and \\ 
Vieri Giuliano Santucci
\affiliations
Institute of Cognitive Science and Technologies, National Research Council, Italy \\
\emails
\{angelo.oddi, riccardo.rasconi, emilio.cartoni, gabriele.sartor, gianluca.baldassarre, vieri.santucci\}@istc.cnr.it
}

\begin{document}

\maketitle

\begin{abstract}

In symbolic planning systems, the knowledge on the domain is commonly provided by an expert. Recently, an automatic abstraction procedure has been proposed in the literature to create a Planning Domain Definition Language (PDDL) representation, which is the most widely used input format for most off-the-shelf automated planners, starting from `options', a data structure used to represent actions within the hierarchical reinforcement learning framework. We propose an architecture that potentially removes the need for human intervention. In particular, the architecture first acquires options in a fully autonomous fashion on the basis of open-ended learning, then builds a PDDL domain based on symbols and operators that can be used to accomplish user-defined goals through a standard PDDL planner.
 We start from an implementation of the above mentioned procedure tested on a set of benchmark domains in which a humanoid robot can change the state of some objects through direct interaction with the environment. We then investigate some critical aspects of the information abstraction process that have been observed, and propose an extension that mitigates such criticalities, in particular by analysing the type of classifiers that allow a suitable grounding of symbols.

\end{abstract}

\section{Introduction}
One of the main challenges in Artificial Intelligence is the problem of abstracting high-level models directly leveraging the interaction between the agent and the environment, where such interaction is typically performed at low-level through the agent’s sensing and actuating capabilities.
Such information abstraction process indeed reveals invaluable for high-level planning, as it allows to make explicit the causal relations existing in the abstracted model which would otherwise remain hidden at low-level.
In this respect, some interesting work has been done in the recent literature. For instance, in ~\cite{konidaris2018skills} an algorithm is presented for automatically producing symbolic domains based on the Planning Domain Definition Language (PDDL, see~\cite{Ghallab98}), also trying to build hierarchical abstractions in ~\cite{konidaris-skills-loop}, starting from a set of low-level skills represented in the form of \textit{abstract subgoal options}.
The contribution of this work is twofold.
On the one hand, we extend the scope of the information abstraction procedure proposed in~\cite{konidaris2018skills} by directly linking the latter with a goal-discovering and skill-learning robotic architecture (GRAIL), see~\cite{DBLP:journals/tamd/SantucciBM16}, capable of autonomously producing a set of low-level skills through intrinsically motivated learning ~\cite{oudeyer2007intrinsic,baldassarre2013Book}.
Such skills will then be used as input for the subsequent abstraction process, thus creating an automated information processing pipeline from the low-level direct interaction of the agent with the environment, to the corresponding high-level PDDL domain representation of the same environment.
On the other hand, given a set of selected low-level domains in which GRAIL is set to operate, we carry on an analysis on the features of the abstracted PDDL representations depending on the categorization capabilities of the classifiers used for the production of the symbolic vocabulary, thus shedding some light on a number of interesting correlations between low-level generalization capabilities of the abstraction procedure and the quality of the produced high-level domains, sketching some guidelines on how this information can be used to increase the completeness  of the obtained PDDL domains, as well as the efficacy of autonomous environmental knowledge acquisition.

The paper is organized as follows. Section~\ref{sec:GRAIL} will briefly describe the GRAIL system; Section~\ref{sec:pddlgen} will summarize the features of the information abstraction procedure and Section~\ref{sec:integration} will provide some empirical insights stemming from the integration of the previous two systems. Finally, Section~\ref{sec:conclusions} will end the paper with some concluding remarks.

\section{The GRAIL Skill Learning System}
\label{sec:GRAIL}
We decided to apply the abstraction procedure on the output of M-GRAIL,
an advancement of a previous architecture called GRAIL (Goal-Discovering Robotic Architecture for Intrinsically-Motivated Learning, \cite{DBLP:journals/tamd/SantucciBM16}) that in turn is the result of a series of increasingly more complex systems \cite{Santucci2013iccm,Santucci2014icdl}. 
GRAIL is an open-ended learning system that 
discovers new interesting events while interacting with the environment and store them as ``goals''. 
GRAIL then automatically train itself through \textit{intrinsic motivation} to achieve these goals from different starting conditions.
For each goal, GRAIL builds a separate "skill" that achieves that goal.
By using competence-based intrinsic motivation GRAIL focuses its training to achieve the highest overall competence (i.e. reliability) on all skills as fast as possible.
M-GRAIL \cite{Santucci2019autonomous} also keeps a series of predictors that predict the percentage of success  of the skill depending on the starting condition, thus enabling M-GRAIL to recognize when the skill can be successfully initiated.

\section{The Information Abstraction Procedure}
    \label{sec:pddlgen}
The information abstraction procedure (called \textit{\textbf{PDDL-Gen}} in this work) has the objective of transforming the environmental low-level knowledge learned by the agent in a PDDL-based representation of the operational domain suitable for high-level planning.

This section is dedicated to providing the \textit{intuition} behind \textit{\textbf{PDDL-Gen}}, so as to properly pave the way for the subsequent sections; the fully detailed description of the domain abstraction procedure can be found in~\cite{konidaris2018skills}.

In order to help the description we will make use of a running example devised for a specific domain, described as follows.
Let us consider an environment containing six ball-shaped bulbs (labeled $b_1, b_2, b_3, b_4, b_5, b_6$), e.g., equally spaced in a row, that light up if touched.
Let us suppose that the dynamics of the environment impose that $b_1$ lights up whenever it is touched independently from the state of the other bulbs, while bulb $b_i$ lights up when touched only if $b_{i-1}$ is lighted up (enabling precondition), with the exception of $b_6$, which lights up when touched and concurrently switches off all the lighted bulbs.
In addition, we will assume that the agent has already learned all the available skills to interact with the environment by means of the M-GRAIL architecture, described in Section~\ref{sec:GRAIL}.
Such skills are represented through the following options: $o_1 \leftarrow$ light up $b_6$ and switch off all the lighted bulbs, $o_2 \leftarrow$ light up $b_1$, $o_3 \leftarrow$ light up $b_2$, $o_4 \leftarrow$ light up $b_3$, $o_5 \leftarrow$ light up $b_4$, and $o_6 \leftarrow$ light up $b_5$.   
\textit{\textbf{PDDL-Gen}} proceeds according to following steps.

\paragraph{Computing the options' characterizing sets.} The procedure is supposed to accept in input an option-based (see~\cite{Sutton:1999:MSF:319103.319108}) representation for each skill previously learned by the agent, expressed in the form of two classifiers for each option (a.k.a. the option's \textit{characterizing set}), namely the \textit{Initiation Set} classifier $Cl(I)$ and the \textit{Effect Set} classifier $Cl(E)$.
The training process for both classifiers will be described in Section~\ref{sec:integration}.

\begin{figure}[ht]
    \centering
    \includegraphics[scale=0.3]{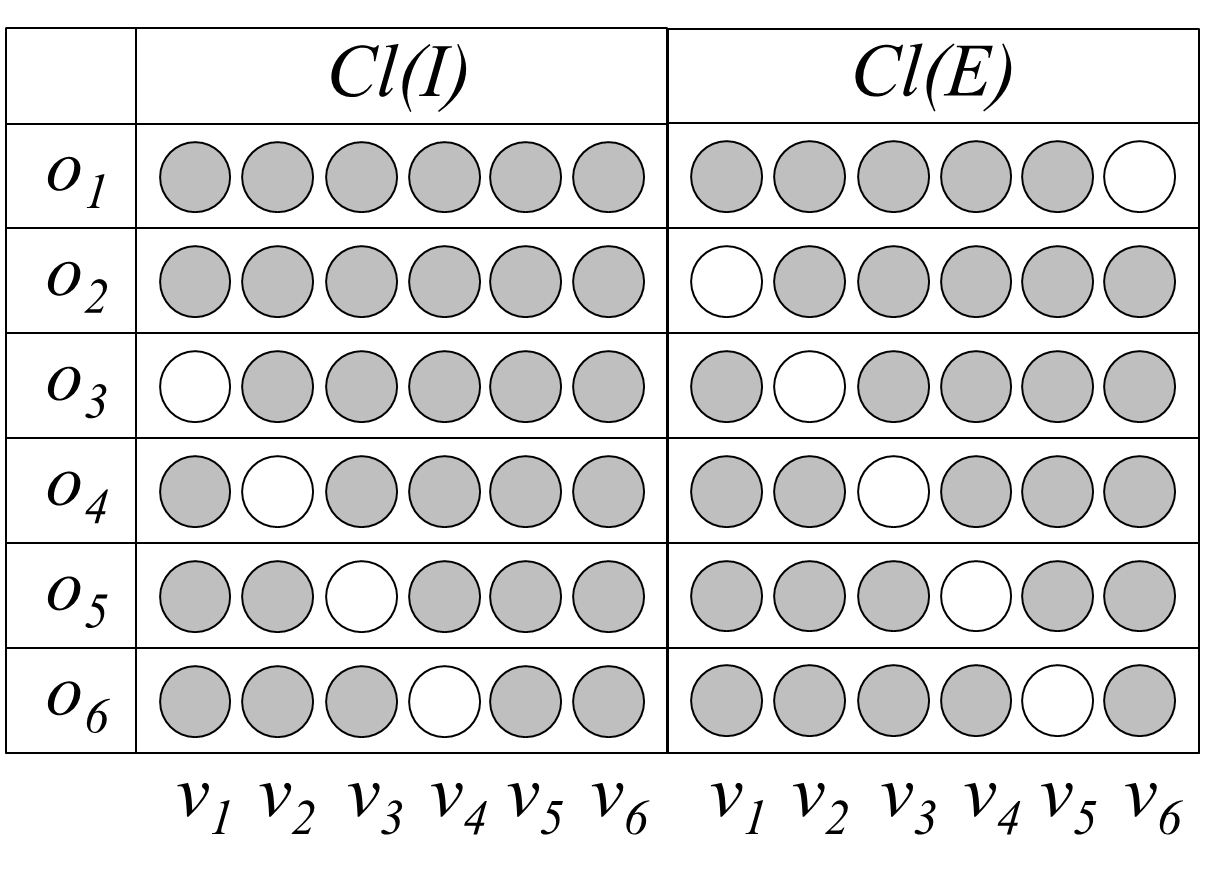}
    \caption{Running example: C4.5 classifiers}
    \label{fig:scenario_1_J48_classifiers}
\end{figure}

Figure~\ref{fig:scenario_1_J48_classifiers} presents a graphical representation of the two $Cl(I)$ and $Cl(E)$ classifiers for each option, trained from the data obtained through the agent's interactions with the six-bulbs environment described above\footnote{The classifiers are computed using the WEKA toolkit (\cite{hall09}) C4.5 decision tree algorithm (\cite{Quinlan:1993:CPM:152181}).}. 
For example, let us look at the $o_4$ row, $Cl(I)$ column in the figure, representing the Initiation Set classifier of option $o_4$.
For the sake of simplicity, we make the assumption that each bulb $b_i$ is represented by a single low-level variable $v_i$; in general, a white-colored $v_i$ represents $v_i = ON$, a black-colored $v_i$ represents $v_i = OFF$, and a grey-colored $v_i$ conveys the information that $v_i = $ \textit{don't care} (i.e., its value is unimportant for classification).
To wrap-up, $Cl(I)$ will classify as belonging to the Initiation Set of $o_4$ only those low-level states $s = \{v_1, v_2, v_3, v_4, v_5, v_6\}$ where $v_2 = ON$, irrespectively of the other variables.
Similarly, $Cl(E)$ will classify as belonging to the Effect Set of $o_4$ only those low-level states where $v_3 = ON$.

\paragraph{Computing the factors.} All the options that are taken into account in this work satisfy the so-called \textit{abstract subgoal option} condition, meaning that their execution will only change a specific subset of the available low-level variables $V = \{v_1, v_2, ..., v_n\}$ (i.e., the option's \textit{mask}), leaving the remaining ones unaltered.
As a consequence of this feature, the whole set of low-level variables $V$ can be \textit{factorized} in $m$ subsets $F = \{f_1, f_2, ..., f_m\}$ called \textit{factors}, such that $f_i \cap f_j = \emptyset$ for any $f_i, f_j \in F$ ($i \neq j$), and $V = f_1 \cup f_2 \cup ... \cup f_m$.
More specifically, each factor $f_i$ returned by the factorization process \textit{is the collection of all the state variables modified by the same set of options}.
The product of the factorization process (i.e., the \textit{factors} thus defined) represent an essential element for the subsequent step of the information abstraction procedure, that is, the synthesis of the symbolic vocabulary.

\begin{figure}[ht]
    \centering
    \includegraphics[scale=0.3]{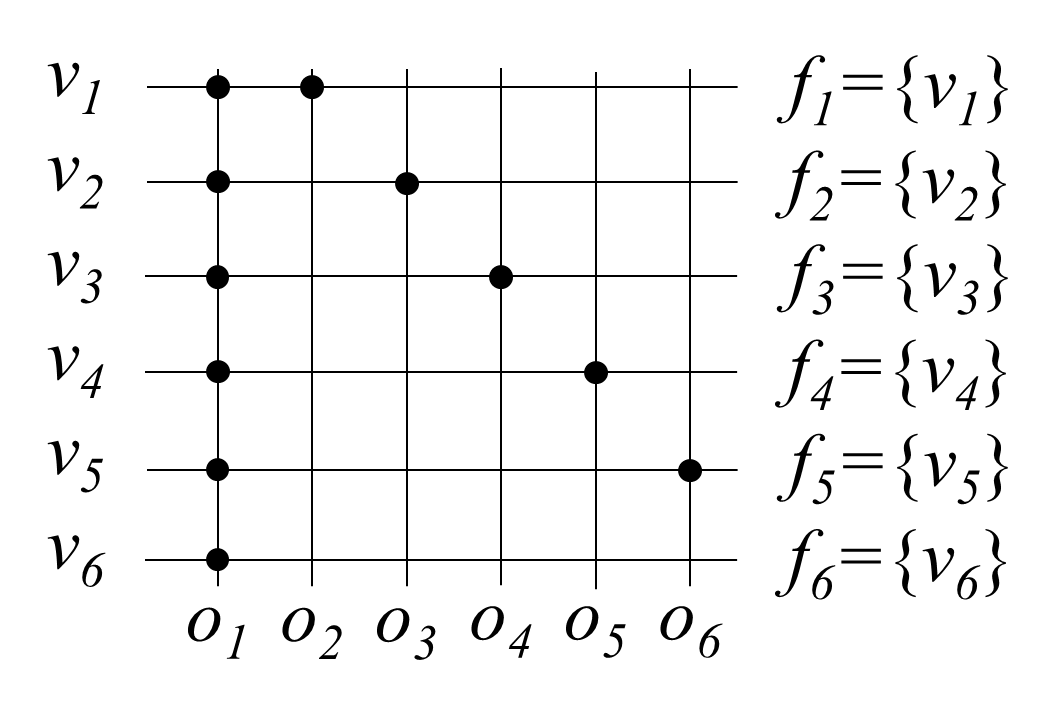}
    \caption{Running example: factors computation}
    \label{fig:scenario_1_J48_factors}
\end{figure}

Continuing the previous example, Figure~\ref{fig:scenario_1_J48_factors} presents the factors that are obtained considering the options' characterizing sets in the bulbs domain.
The figure shows a grid in which the $y$ axis contains the low-level variables, the $x$ axis contains the options, and each dot in the intersection $\langle v_i, o_j \rangle$ represents the fact that state variable $v_i$ is modified by the option $o_j$.
The list of obtained factors $\{f_1, f_2, f_3, f_4, f_5, f_6\}$ is depicted on the right side of the figure.
Note that each factor contains one variable only, as each single variable is modified by a different subset of options. 
  
\paragraph{Generating the symbol set.} Given the set of options $O = \{o_1, o_2, ..., o_r\}$ provided in input, the objective of this step is to produce the \textit{complete} symbolic vocabulary (let us name it $P$, initially empty).

The symbol generation phase proceeds as follows.
For each option $o_i \in O$, the set of factors $factors(o_i)$ containing the variables modified by $o_i$ is computed, as well as $o_i$'s effect set $\textit{Effect}(o_i)$ (i.e., the set of low-level states that the agent can possibly reach after executing $o_i$).

Then, the procedure enters the symbol production cycle, in which every factor $f_j \in factors(o_i)$ is tested for independence in $\textit{Effect}(o_i)$ ($f_j$ is independent in the effects of $o_i$ if the values taken by the variables $v_h \in f_j$ do not depend by any other variable $v_i \in f\setminus f_j$, within the scope of $o_i$'s effect set subspace).
This test is very important for the correct production of symbols, as any factor $f_j$ that is independent in $\textit{Effect}(o_i)$ can be turned into \textit{one single symbol} $\sigma_j$ that represents all the variables contained in $f_j$, safely disregarding the other variables in $\textit{Effect}(o_i)$, as $o_i$ modifies the variables in $f_j$ always as a single block.
Conversely, provided that $f$ is the set of factors that did not pass the independence test, it is necessary to produce a different symbol for any subset of factors $ f_s \subset f$.

\begin{figure}[ht]
    \centering
    \includegraphics[scale=0.68]{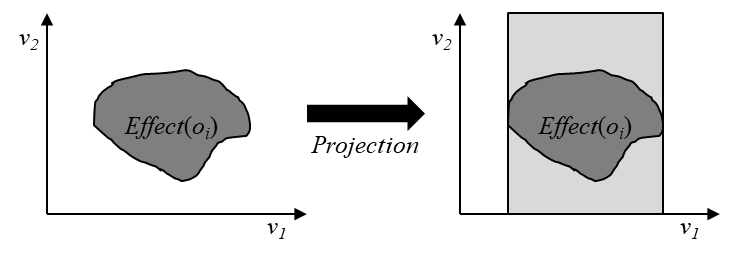}
    \caption{A visualization of the projection operation. Considering the $\textit{Effect}(o_i)$ on the left side of the figure, with $f = \{f_1, f_2\}$ where $f_1 = \{v_1\}$ and $f_2 = \{v_2\}$, the projection of $f \setminus f_1 = \{v_2\}$ out of $\textit{Effect}(o_i)$ removes the restrictions based on the state variables in $f \setminus f_1$, resulting in the light-grey set on the right side.}
    \label{fig:projEx}
\end{figure}

Each produced symbol $\sigma_j$ is characterized by a label (defining its name) and a new classifier $Cl(\sigma_j)$ whose task is to classify the set of low-level state for which $\sigma_j$ is verified.
The computation of $Cl(\sigma_j)$ is of paramount importance, and proceeds by \textit{projecting out} from $\textit{Effect}(o_i)$ all the low-level variables $v \in f\setminus f_j$ (see Figure~\ref{fig:projEx}); hence, $Cl(\sigma_j)$ is ultimately the classifier that discriminates the low-level state set resulting from the previous projection.

\begin{figure}[ht]
    \centering
    \includegraphics[scale=0.30]{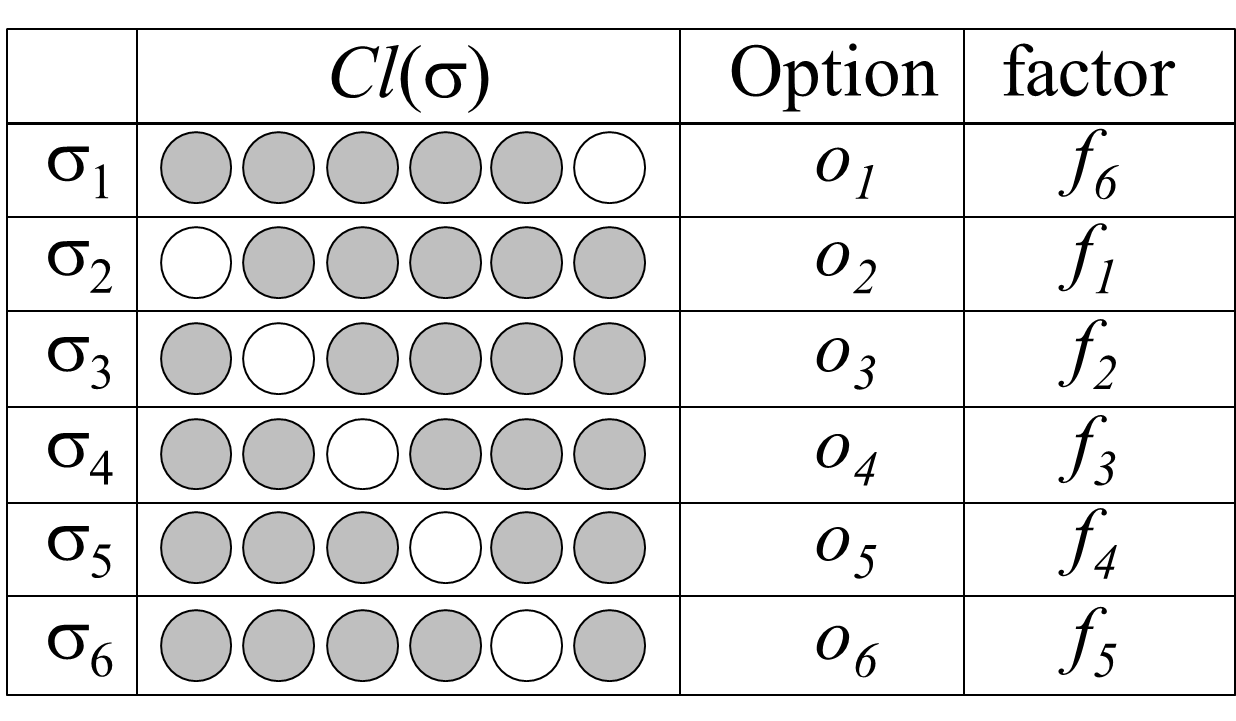}
    \caption{Running example: the produced symbols}
    \label{fig:symbols}
\end{figure}

The table in Figure~\ref{fig:symbols} presents the list of the produced symbols $\sigma_i \in \{\sigma_1, \sigma_2, \sigma_3, \sigma_4, \sigma_5, \sigma_6\}$ relatively to our bulbs domain, where the $i$-th row of the table describes the symbol $\sigma_i$.
In particular, the \textit{Cl($\sigma$)} column shows the \textit{grounding classifier}\footnote{Grounding classifiers discriminate the set of low-level states in which the symbol holds, thus providing the symbol's \textit{semantics}.} associated with $\sigma_i$ (according to the same convention used in Figure~\ref{fig:scenario_1_J48_classifiers}), the \textit{Option} column shows the option that has produced $\sigma_i$ as one of its effects, and the $factors(\sigma)$ column shows the list of factors over which $\sigma_i$'s grounding classifier is defined.
Note that due to the simplicity of the selected example, each produced symbol in the figure is identical to one of the effect set classifiers $Cl(E)$ in Figure~\ref{fig:scenario_1_J48_classifiers}.
In general though, the symbol generation procedure returns symbol sets whose grounding classifiers can significantly differ from $Cl(E)$, thus symbolically abstracting the relevant aspects of reality at a finer level of granularity.


\paragraph{Generating the PDDL operator descriptions.}
Once the complete set of symbols $P$ has been created, it is possible to express our model as a set-theoretic high-level domain specification using the Planning and Domain Definition Language (PDDL) formalization (\cite{Ghallab98}), which is the most widely used input format for most off-the-shelf automated planners.

A set-theoretic specification is expressed in terms of a set of propositional symbols $P = \{\sigma_1, ..., \sigma_n\}$ (each associated to a grounding classifier $Cl(\sigma_i)$) and a set of operators $A = \{op_1, ..., op_m\}$.
Each operator $op_i$ is described by the tuple $op_i = \langle pre_i, \textit{eff}^+_i, \textit{eff}^-_i)$, where $pre_i$ contains all the propositional symbols that must be $true$ in a state $s$ for $opt_i$ to be executed from $s$, while $\textit{eff}^+_i$ and $\textit{eff}^-_i$ contain the propositional symbols that are respectively set to $true$ or $false$ after $op_i$'s execution. All the other propositional symbols remain unaffected by the execution of the operator.

In order to produce a correct PDDL representation, it is therefore necessary to populate the three sets ($pre_i$, $\textit{eff}^+_i$ and $\textit{eff}^-_i$) \textit{for each option} $o_i$ by properly selecting which symbols, among those contained in $P$, will fall in any of such sets.

\textit{\textbf{Effects computation}}.
With the previous assumptions, all the symbols $\sigma_j \in P$ that are produced as an effect of $o_i$ (see the symbol generation process) will become part of $\textit{eff}^+_i$ (i.e., the option's \textit{direct effects}).

Contextually, all the symbols $\sigma_j \in P$ that are not produced as an effect of $o_i$ (see the symbol generation process) and whose factors are entirely contained in $factors(o_i)$, will become part of $\textit{eff}^-_i$, as their truth value is modified by $o_i$'s execution (\textit{full overwrites}).

For the same reason, all the symbols $\sigma_j \in P$ whose factors are \textit{partially} contained in $factors(o_i)$ will also become part of $\textit{eff}^-_i$ (\textit{partial overwrites}); but in order to correctly identify the symbolic element unmodified by $o_i$'s execution, it is necessary to set to $true$ the symbol $\sigma_k \in P$ defined by projecting out of $\sigma_j$ all the variables modified by $o_i$. Consequently, $\sigma_k$ will become part of $\textit{eff}^+_i$.

Obviously, all the symbols $\sigma_j \in P$ whose factors are entirely out of $factors(o_i)$ (i.e., that are not part of $o_i$'s effects) will remain $true$.

\textit{\textbf{Preconditions computation}}.
Option $o_i$'s preconditions are calculated as the union of all the subsets of symbols $P_c \subseteq P$ such that the following conditions hold. 
\begin{enumerate}
\item The union of all the $factors(\sigma)$ related to all symbols $\sigma \in P_c$ must be contained in the set of factors over which $o_i$'s initiation set classifier $Cl(I)$ is defined: $\cup_{\sigma \in P_c} factors(\sigma) \subseteq factors(Cl(I))$.
    
\item The intersection of all the grounding classifiers related to the symbols $\sigma \in P_c$ (i.e., the logical \textit{and} among such symbols) must return a set of states that is a subset of $o_i$'s initiation set: $\cap_{\sigma \in P_c} Cl(\sigma)\subseteq Cl(I)$.
    
\item Since, according to the presented model, no two symbols with grounding classifiers defined over the same factors can be \textit{true} at the same time, it is also necessary to guarantee that no two symbols contained in $P_c$ have any factor in common: $factors(\sigma_i) \cap factors(\sigma_j) = \emptyset$, for each  $\sigma_i, \sigma_j \in P_c$ and $\sigma_i \neq \sigma_j$.
\end{enumerate}

\begin{figure}[h]
    \centering
    \includegraphics[scale=0.35]{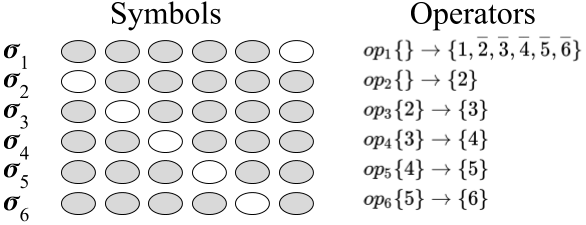}
    \caption{\textit{Reset} PDDL domain using \textbf{\textit{C4.5}}}
    \label{fig:scenario_Reset_J48}
\end{figure}

The complete symbolic representation of the example domain is presented in Figure~\ref{fig:scenario_Reset_J48}.
The \textit{Operators} column lists the PDDL equivalent of each option, expressed in the format $op_i \{\sigma \mid \sigma \in pre_i\} \rightarrow \{\sigma \mid \sigma \in eff^+_i\} \cup \{\overline{\sigma} \mid \sigma \in eff^-_i\}$.
Note that all symbols $\sigma_i$ used to characterize the operators are represented using their index $i$ only ($i \equiv \sigma_i$ and $\overline{i} \equiv \overline{\sigma_i}$).

\section{Empirical analysis of System Integration}
\label{sec:integration}

\subsection{Choosing a learner}
We chose GRAIL because it learns abstract subgoal option-like skills, i.e. modules that once activated perform motor activities that reliably lead to a particular "goal" i.e. some variables of the world staying within a certain range (just as abstract subgoal option lead to a termination condition where a subset of variables will stay in some set of values regardless of the starting condition).

\subsection{Building the datasets for PDDL-Gen}
To build the datasets needed for the classifiers and the set representations of the initiation and effects set, we chose to use data from each skill only after that skill had become fully reliable (i.e. it no longer fails); this assumes non-stochastic environment where it is possible to learn skills with guaranteed success. 
To build the $Cl(I)$ classifier training dataset, we considered as positive cases all the low-level variable values before the successful execution of the skill, and as negative cases all the low-level variable values in conditions where GRAIL has tried to execute the skill but the predictor has always been zero.
To build the $Cl(E)$ classifier training dataset, we considered as positive cases all the low-level variable values after the skill has been successfully executed. As negative effect cases, we used all the low-level variable values before the execution of that skill, whether it succeeded or not (since we know that GRAIL will not execute the skill if its goal/effect is already achieved).
As for the masks dataset, a collection of all successful executions of the skill was used, to compare the variables before and after the execution and see which ones are affected by each skill.


\subsection{Choosing a classifier and turn it into a compact set representation}
\label{subsec:classifierSelection}
PDDL-Gen requires that the projection operator is applied to the initiation and effect sets.
So it is important that the initiation and effect sets are represented with a data structure that lends itself to be "projected", as exemplified in Figure~\ref{fig:projEx}.
However, PDDL-Gen does not explicitly state the data structure on which this operator can be applied, thus leaving such choice as an implementation decision.

As a matter of fact, not all classifiers offer a set representation that is easily projected.
As an example, deep neural networks might offer good classification performances, but their classification is encoded in thousands of neural weights, a data structure which does not readily offer a way to construct a projection.
On the other hand, classifiers that builds a decision tree, such as C4.5 (used in \cite{konidaris2018skills}), can be easily converted into a "projectable" set representation.
In particular, we can build a representation of a set from a decision tree as a series of filters on each variable (see below). This set representation can be easily projected by simply removing all filters for the projected variables.

\begin{figure}[ht]
    \centering
    \includegraphics[scale=0.40]{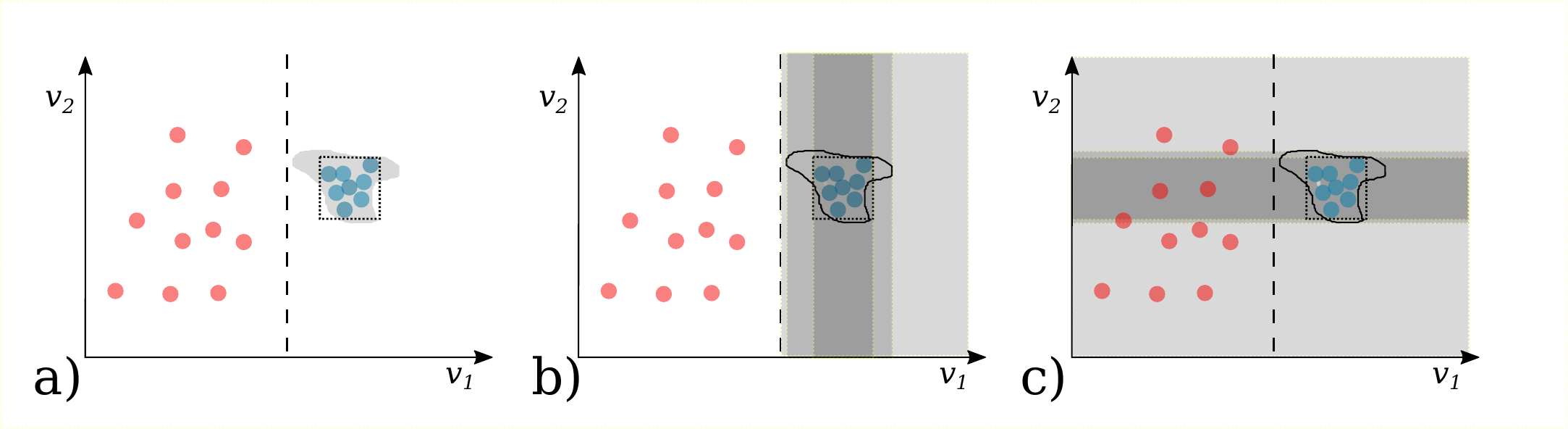}
    \caption{Projections on three different set representations -  a) blue circles represent positive examples, red circles are negative examples; the shape in grey is the "true" set representation as given by an oracle, the \textbf{"IntM"} representation is shown as a dotted box and the one from a classifier such as C4.5 as a dashed line; b) and c) show the resulting projections of the three representations on the two factors,  $f_1 = \{v_1\}$ and $f_2 = \{v_2\}$).}
    \label{fig:projection}
\end{figure}

However, building such a representation from a C4.5 decision tree, does not always yield optimal results.
As shown in Figure \ref{fig:projection}, the filters derived from a C4.5 decision tree will try to optimize the discrimination capability, however this might result in a set representation that is considerably larger than what the data suggest and it may even lack constraints on some of the factors of the mask.
We will show in the scenarios below how this can negatively impact the PDDL-Gen.
To amend this problem, we have developed a method to derive a projectable set representation that compactly describes the initiation and effect set. We will call this method ``\textit{Intersection+Mask}'' (\textbf{\textit{IntM}}), and compare it to the simpler representation obtained through \textbf{\textit{C4.5}}.
The two set representations are obtained as follows:
\begin{itemize}
    \item \textbf{C4.5} - The set representation of $I(o_i)$ and $\textit{Effect}(o_i)$ are derived from the \textit{decision trees}, generated with the C4.5 algorithm on the respective datasets.
    In particular, for each \textit{true} leaf of the decision tree a compact set representation is built. Each compact set representation is a collection of filters reflecting all the decisions to reach that leaf: for all variables used as decision point a filter is added so that only values which would have passed the decision point are retained (i.e. a decision point which states $v_1 > 0.7$ on a variable which goes from 0 to 1 generates a filter $0.7 < v_1 < 1$. 
    All filters are then joined together by logical "AND".
    If multiple true leaves exists, the compact set representations are joined together into a single set representation by logical "OR".
    This set representation can be easily projected by simply removing the filters whose variables belong to the factor that is being projected.
    
    \item \textbf{IntM} - As in C4.5, for each \textit{true} leaf of the decision tree a compact set representation is built.
    However, the filter values are not taken from the decision tree but are generated by looking at the values of all positive examples belonging to that leaf. In particular, for each variable a filter is built that only accepts values between the lowest and the highest values of the variable found in the positive examples.
    In the case of the $\textit{Effect}(o_i)$, only variables belonging to the mask are used, while for $I(o_i)$ all variables are used.
\end{itemize}

In this work we will restrict ourselves to scenarios where decision trees will have one \textit{true} leaf only, ruling out the possibility of disjunctive preconditions and/or effects.

Note: as we pointed out above, not all variables are necessarily used as decision points, so the C4.5 set representation might be defined over less variables than the \textbf{IntM} one, which will instead provide tighter bounds around the set (see Figure \ref{fig:projection}).

 In the following empirical analysis, we will see how the choice of the classifier to represent both the set $I(o_i)$ and the $\textit{Effect}(o_i)$ sets affects the output PDDL representation and its correctness in a series of test scenarios of the \textit{bulbs} domain above introduced.

\subsection{Empirical analysis}
In this section we analyze a number of relevant features in the representations obtained using the \textit{C4.5}, and \textit{IntM} classifiers, testing them on three different scenarios: (i) the previously introduced running example (henceforth referred to as  \textit{Reset} scenario), (ii) a scenario where the addition of some negative effects to the output PDDL representation depends on the kind of classifier used (\textit{Negative} scenario), and (iii) a scenario where some states cannot be reached by the robot actions (\textit{Unreachable} scenario).

\subsubsection{\textit{Reset} scenario.}
The list of available options that characterize this domain, as well as the description of its dynamics, have been presented in Section~\ref{sec:pddlgen}.

\begin{figure}[ht]
    \centering
    \includegraphics[scale=0.32]{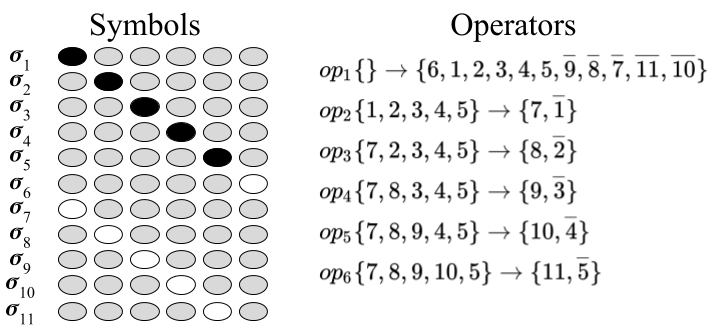}
    \caption{\textit{Reset} PDDL domain using \textbf{\textit{IntM}}}
    \label{fig:scenario_Reset_int_mask}
\end{figure}

The symbolic abstraction of this scenario returned by the \textit{PDDL-Gen} procedure using the \textbf{\textit{C4.5}} and the \textbf{\textit{IntM}} classifiers are respectively shown in Figure~\ref{fig:scenario_Reset_J48} and Figure~\ref{fig:scenario_Reset_int_mask}.
Understandably, the characteristics of the symbol sets obtained are caused by the different classifications features (outlined in the previous section) of the used classifiers.
In particular, we observe that while both classifiers produce common symbols (i.e., $\{\sigma_1, ..., \sigma_6\}$ from the \textbf{\textit{C4.5}} case and $\{\sigma_6, ..., \sigma_{11}\}$ from the \textbf{\textit{IntM}} case), the latter classifier produces $5$ more symbols through which it is possible to define the \textit{off} state of each individual bulb ($\{\sigma_1, ..., \sigma_5\}$).

Interestingly, from the description of the scenario dynamics (see in particular the option $o_1$ that switches off all the bulbs), a \textit{complete} PDDL representation would be one that contains the necessary symbols to represent the \textit{off} state of each individual bulb.
If we analyze the PDDL domains obtained with both classifiers, we observe that such symbols are only obtained in the \textbf{\textit{IntM}} case (symbols $\sigma_1, ..., \sigma_5$).
The presence of the previous symbols has important consequences on the representation capability of the obtained PDDL, as it allows to define the low-level state in which the $b_1, ..., b_5$ bulbs are \textit{off}.
Conversely, through the \textbf{\textit{C4.5}}-based PDDL, it is impossible to explicitly represent such state though it is a state in which the agent may find itself during the execution of a plan\footnote{Set-theoretic PDDL forbids the use of negative preconditions.}!

Despite this limitation, both classifiers produce syntactically correct PDDL domains that can be used for automated planning, as can be easily verified by testing the domains on problem instances built with the obtained symbols.

\subsubsection{Scenario \textit{Negative}}
The dynamics of this scenario is similar to the previous case, except: (i) touching $b_1$ lights up $b_1$ \textit{and} $b_2$, (ii) $b_2$ lights up whenever it is touched independently from the state of the other bulbs, and (iii) $b_6$ lights up if touched only if $b_5$ is \textit{on}.
The skills the agent has learned to operate in this scenario are represented through the following options: $o_1 \leftarrow$ light up $b_2$, $o_2 \leftarrow$ light up $b_3$, $o_3 \leftarrow$ light up $b_1$ and $b_2$, $o_4 \leftarrow$ light up $b_4$, $o_5 \leftarrow$ light up $b_5$, and $o_6 \leftarrow$ light up $b_6$.


\begin{figure}[ht]
    \centering
    \includegraphics[scale=0.37]{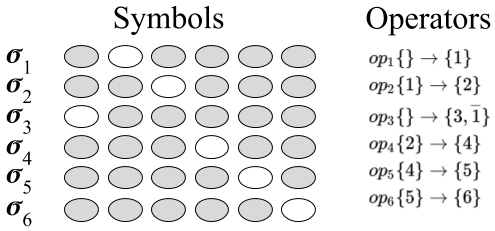}
    \caption{Scenario \textit{Negative}: C4.5 classifier}
    \label{fig:scenario_negative_c45}
\end{figure}


The PDDL symbols and operators obtained from PDDL-Gen using the \textbf{\textit{C4.5}} classifier is shown in Figure~\ref{fig:scenario_negative_c45}.
As a remarkable aspect of the returned \textbf{\textit{C4.5}} domain, we immediately observe that operator $op_3$ has (correctly) no preconditions, adds symbol $\sigma_3$ as positive effect (i.e., bulb $b_1$ \textit{on}), but surprisingly includes symbol $\sigma_1$ as negative effect (i.e., switches $b_2$ \textit{off}), \textit{while} $\sigma_1$ \textit{should be included among the positive effects}!

The reason why $\sigma_1$ is not added to the positive effects can be explained by considering the symbol generation process described in Section~\ref{sec:pddlgen}, together with the \textbf{\textit{C4.5}}'s classification features discussed in Section~\ref{subsec:classifierSelection}.
Specifically, about option $o_3$ we note the following: (i) it modifies $b_1$ and $b_2$ ($mask(o_3) = \{v_1, v_2\}$; its initiation set classifier $Cl(I)$ is \textit{empty} ($o_3$ has no preconditions); (iii) its \textbf{\textit{C4.5}}-based effect set classifier discriminates as $o_3$’s effects \textit{only} all the low-level states where $b_1$ is \textit{on}, disregarding the state of $b_2$.
In other words, despite both $b_1$ and $b_2$ are always switched on by $o_3$ ($factors(o_3) = \{f_1, f_2\}$, where $f_1 = \{b_1\}$ and $f_2 = \{b_2\}$), the C4.5 classifier represents the set $\textit{Effect}(o_3)$ only on the basis of the factor $f_1 = \{b_1\}$. Hence, only the symbol $\sigma_3$ (representing $b_1$ on) is generated by $o_3$, and added as positive effect.
Moreover, since symbol $\sigma_1$ (generated by option, $o_1$) satisfies the relation $factors(\sigma_1) = \{f_1\} \subseteq factors(o_3) = \{f_1, f_2\}$ (see the effect computation process described in Section~\ref{sec:pddlgen}), it is included as negative effect of option $o_3$.




Conversely, the \textit{IntM} classifier produces correct results: both symbols $\sigma_1$ and $\sigma_3$ are included as positive effects of $opt_3$ (the figure is not shown for reasons of space).
In fact, in this case we have a different representation of the effect set classifier $\textit{Effect}(o_3)$, such that it accepts all the low-level states where both $b_1$ and $b_2$ are \textit{on}.


\subsubsection{Scenario \textit{Unreachable}}
The dynamics of this scenario is similar to the previous case, except: (i) $b_1$ lights up only if $b_2$ is \textit{off} (enabling precondition), (ii) $b_2$ lights up whenever it is touched independently from the state of the other bulbs, and (iii) the bulb $b_6$ is ineffective (no reset).
The bulbs are periodically set to \textit{off} by the environment, but the agent has no way to reset them.
The skills the agent has learned to operate in this scenario are represented through the following options: $o_1 \leftarrow$ light up $b_2$, $o_2 \leftarrow$ light up $b_3$, $o_3 \leftarrow$ light up $b_1$, $o_4 \leftarrow$ light up $b_4$, and $o_5 \leftarrow$ light up $b_5$.

Relatively to this scenario, we see that both the \textbf{\textit{C4.5}} and the \textbf{\textit{IntM}} classifiers produce exactly the same set of symbols, each symbol defining the \textit{on} status of each bulb, irrespective of all the other bulbs.
Yet, the PDDL abstraction returned by the classifiers is different, due to the differences existing between their respective \textit{characterizing sets}.
For example, we observe that in the \textbf{\textit{C4.5}} case, the operator $op_4$ only requires that $b_3$ is \textit{on} ($\sigma_2$) as a precondition for lighting up $b_4$, while in the \textbf{\textit{IntM}} case, the same operator requires that also $b_2$ is \textit{on} ($\sigma_1$). 
Though this precondition for lighting up $b_4$ is not required by the dynamics of the Unreach scenario, the reason why it is introduced lies in the different discrimination capability of the \textbf{\textit{C4.5}} w.r.t. the \textbf{\textit{IntM}} classifier, as the former tends to minimize the number of necessary variables for classification, as described in Section~\ref{subsec:classifierSelection}.




The interesting aspect of this scenario is that no classifier succeeds in capturing option $o_3$'s enabling condition, requiring $b_2$ to be \textit{off} in order for $b_1$ to lighted up.  
This circumstance can be readily explained by the fact that this scenario contains no options that switch off the bulbs.
Hence, as PDDL-Gen only generates symbols from the effects, the necessary symbol that expresses the concept of \textit{$i$-th bulb is off} can never be obtained, regardless the type of classifier used.

\begin{figure}[ht]
    \centering
    \includegraphics[scale=0.37]{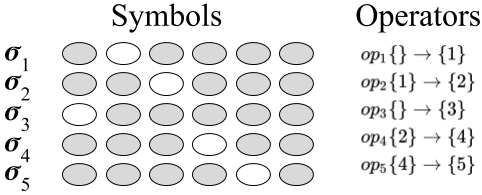}
    \caption{Scenario \textit{Unreachable}: \textit{C4.5} classifier}
    \label{fig:scenario_reach_c45}
\end{figure}

\begin{figure}[ht]
    \centering
    \includegraphics[scale=0.37]{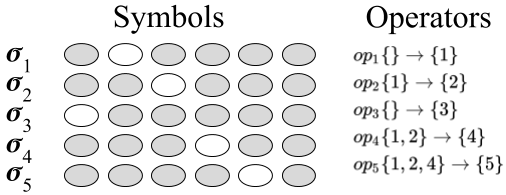}
    \caption{Scenario \textit{Unreachable}: \textit{IntM} classifier}
    \label{fig:scenario_reach_int_mask}
\end{figure}


\section{Conclusions}
\label{sec:conclusions}
In this paper we have connected a goal-discovering and skill-learning robotic architecture (GRAIL) see~\cite{DBLP:journals/tamd/SantucciBM16} to the abstraction procedure proposed in~\cite{konidaris2018skills}, creating a processing pipeline from the low-level direct interaction of the agent with the environment, to the corresponding symbolic representation of the same environment. Subsequently, we have tested the ability of the given abstraction procedure to construct a symbolic representation starting from the agent's learned options. We have carried on a empirical analysis on a number of interesting correlations between low-level generalization capabilities of the abstraction procedure and the completeness/quality of the produced high-level symbolic domains.
Among the possible directions of future work we consider: (i) extend the proposed analysis to the case of abstract sub-goal options with disjunctive preconditions; (ii) the integration of symbolic planning and open-ended learning to increase the ability on one agent to autonomously acquire new skills.


\section*{Achowledgements}
This research has been supported by the European Space Agency (ESA) under contract No. 4000124068/18/NL/CRS, project IMPACT - Intrinsically Motivated Planning Architecture for Curiosity-driven roboTs - and the European Union's Horizon 2020 Research and Innovation Programme under Grant Agreement No 713010, Project ``GOAL-Robots -- Goal-based Open-ended Autonomous Learning Robots''. The view expressed in this paper can in no way be taken to reflect the official opinion of the European Space Agency.

\bibliographystyle{named}
\bibliography{ijcai19}

\end{document}